\documentclass[11pt]{article}
\usepackage{algorithm}
\usepackage{algorithmicx}
\usepackage{algpseudocode}
\usepackage{amsmath}
\usepackage{booktabs}
\usepackage{multirow}
\usepackage{ACL}
\usepackage{placeins}  % 防止表格漂移
\usepackage{array}  % 让表格列宽自定义
\usepackage{caption}
\usepackage{times}
\usepackage{latexsym}
\usepackage{pifont}
\usepackage{pifont}  % 提供 \ding 命令
\usepackage[most]{tcolorbox}
\usepackage{xcolor}
\usepackage[T1]{fontenc}
\usepackage[utf8]{inputenc}
\usepackage{microtype}

\usepackage{inconsolata}

\usepackage{graphicx}

\title{From Implicit Exploration to Structured Reasoning: Leveraging Guideline and Refinement for LLMs}

\author{
  Jiaxiang Chen\textsuperscript{1} \quad
  Zhuo Wang\textsuperscript{1,2} \quad
  Mingxi Zou\textsuperscript{1} \quad
  Zhucong Li\textsuperscript{1} \\
  \textbf{
  Zhijian Zhou\textsuperscript{1,2} \quad
  Song Wang\textsuperscript{3} \quad
  Zenglin Xu\textsuperscript{1}\thanks{~~Corresponding author.} 
  }\\
  \textsuperscript{1}Fudan University \quad
  \textsuperscript{2}Shanghai Innovation Institute \quad
  \textsuperscript{3}Zoom Video Communications\\
  \texttt{jiaxiangchen23@m.fudan.edu.cn}
}

\usepackage{float}

\begin{document}
\maketitle

\begin{abstract}

Large language models (LLMs) have advanced general-purpose reasoning, showing strong performance across diverse tasks. However, existing methods often rely on implicit exploration, where the model follows stochastic and unguided reasoning paths—like walking without a map. This leads to unstable reasoning paths, lack of error correction, and limited learning from past experience.
To address these issues, we propose a framework that shifts from implicit exploration to structured reasoning through \textit{guideline} and \textit{refinement}. First, we extract structured reasoning patterns from successful trajectories and reflective signals from failures. During inference, the model follows these guidelines step-by-step, with refinement applied after each step to correct errors and stabilize the reasoning process.
Experiments on BBH and four additional benchmarks (GSM8K, MATH-500, MBPP, HumanEval) show that our method consistently outperforms strong baselines across diverse reasoning tasks. Structured reasoning with stepwise execution and refinement improves stability and generalization, while guidelines transfer well across domains and flexibly support cross-model collaboration, matching or surpassing supervised fine-tuning in effectiveness and scalability.
%Notably, our structured reasoning approach rivals or surpasses supervised fine-tuning, demonstrating strong scalability and effectiveness.
%Large language models (LLMs) exhibit strong generalization across tasks but often struggle with complex reasoning due to unstable reasoning paths, limited error correction, and lack of experience-based learning. Existing methods typically rely on implicit exploration, leading to inconsistent performance.To overcome these limitations, we propose a framework that transitions from implicit exploration to structured reasoning via two key components: guideline and refinement. We first extract structured reasoning patterns from successful trajectories and reflective signals from failures. At inference time, the model executes these guidelines step by step, applying dynamic refinement after each step to correct errors and enhance stability.Our approach achieves strong and consistent results on the Big-Bench Hard (BBH) benchmark. Detailed analysis shows that stepwise execution improves path stability, refinement enables effective error recovery, and experience-driven learning offers more reliable reasoning guidance.
\end{abstract}
\section{Introduction}

%路径不稳定，缺乏全局约束；

%缺乏失败后的反思与修正机制；

%缺少从经验中学习并生成明确指引的能力，仍依赖隐式自我探索。
%Left: Implicit exploration lacks explicit guidance, often leading to unstable and error-prone reasoning. Right: Structured reasoning leverages a global perspective, enabling more stable execution, self-correction, and learning from prior experience

\begin{figure}[tb]
	\centering
	\includegraphics[scale=0.35]{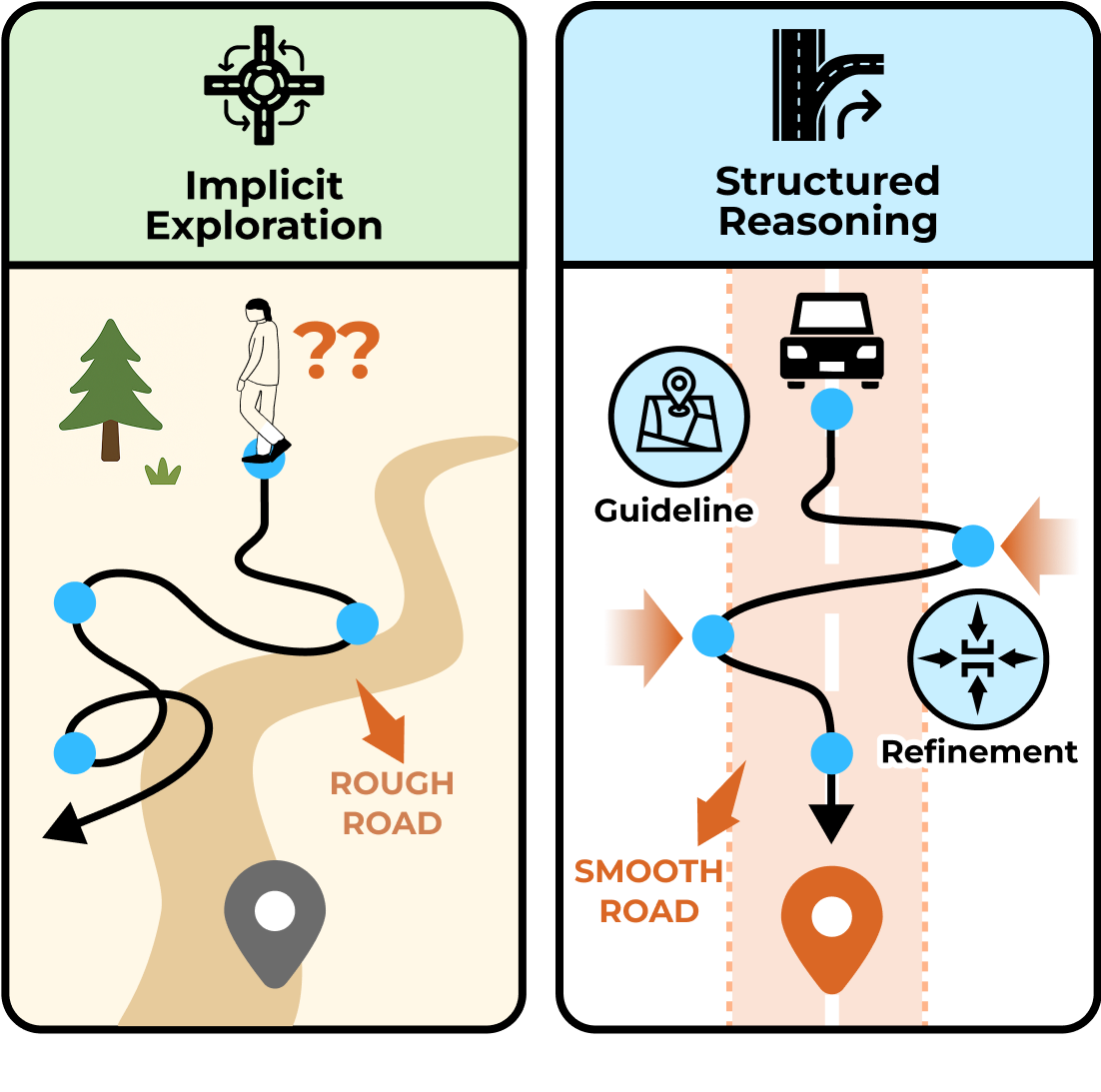}
        \caption{Comparison between \textbf{Implicit Exploration} and \textbf{Structured Reasoning}. \textbf{Left:} Implicit exploration is like walking on a rough path without a map—lacking clear direction, it often leads to unstable and error-prone reasoning. \textbf{Right:} Structured reasoning resembles driving with a roadmap: the \textit{guideline} offers a global route, \textit{refinement} helps correct deviations along the way, and the entire reasoning process remains smoother and more stable—like navigating a wide, well-marked road.}

	\label{fig:illu_cmp}
\end{figure}

Large language models (LLMs)~\cite{gpt4,gpt4o,llama3.1} have demonstrated strong generalization across diverse domains, including mathematics~\cite{mathprompter}, logical reasoning~\cite{logic-lm}, language understanding~\cite{code_understand,utilize_context}, and specialized applications such as finance~\cite{finmem,finhear}. While in-context learning (ICL) enables flexible adaptation, LLMs continue to struggle with complex multi-step reasoning tasks—including planning, symbolic manipulation, and structured decision-making—which are essential for many real-world applications. Overcoming these limitations is critical for improving both the theoretical understanding and practical reliability of LLM-based systems.

Recent work has made progress toward this goal. Many researchers have explored advanced reasoning paradigms built on Chain-of-Thought (CoT)\cite{cot,few-shot_icl,zero-shotcot,sc_cot} prompting. ReAct\cite{react} introduces a "thought–action–observation" cycle that enables interaction with the environment and self-evaluation. Tree-of-Thought (ToT)\cite{tot} represents reasoning as a search over tree-structured trajectories, encouraging exploration of multiple possibilities. Beats~\cite{beats} and FoT\cite{fot} use MCTS\cite{mcts} and voting to improve robustness. Notably, few-shot CoT and FoT attempt to leverage prior examples to improve reasoning. Although these approaches represent progress toward more structured reasoning, they still operate under an implicit reasoning paradigm: they leverage past examples in limited ways, lack global guidance, and offer no principled mechanism for error correction.

Nevertheless, implicit reasoning methods face three persistent challenges in complex multi-step tasks. First, models often struggle to learn from prior experience and extract reusable strategies—reasoning is executed from scratch without leveraging past successes. Second, reasoning trajectories are frequently unstable. In the absence of a well-defined global structure, models tend to deviate from valid reasoning paths, leading to the compounding of early-stage errors. Third, existing methods lack mechanisms for error recovery; once a reasoning path is taken, there is typically no reflection or correction.
As illustrated in Figure~\ref{fig:illu_cmp}, these limitations underscore the need to move from implicit exploration toward structured reasoning guided by learned strategies and iterative refinement.

Therefore, we propose a structured reasoning framework based on guideline extraction and stepwise refinement to address three key challenges in complex reasoning: unstable reasoning paths, lack of error correction, and limited use of learned experience. The framework consists of two stages. First, we extract structured reasoning patterns and key decision points from successful trajectories to form generalizable \textbf{guidelines}. In parallel, we analyze failed cases to identify typical error patterns, which serve as reflective signals. Second, during inference, the model follows the extracted guideline step by step. After each step, a refinement module evaluates the intermediate output and applies targeted corrections. This procedure introduces global planning, enables error recovery, and incorporates experiential learning.

We evaluate our approach on complex reasoning tasks from the Big-Bench Hard (BBH) benchmark~\cite{bbh} across multiple models. Our framework consistently outperforms strong baselines—including CoT, ReAct, ToT, Beats and FoT—achieving notable gains in both accuracy and stability. Further analysis reveals that step-wise execution enhances reasoning coherence through explicit decomposition, refinement enables real-time error correction, and experience-based learning produces effective instructions that outperform implicit self-planning. We also investigate model collaboration in the refinement stage, providing insights into cross-model interactions and division of labor. Additionally, inspired by prior work~\cite{icl_optimize} that frames ICL as implicit optimization, we compare our method to supervised fine-tuning (SFT) using the same model (LLaMA-3.3-70B), and observe better generalization across tasks. These findings suggest that structured reasoning with guideline and refinement not only improves interpretability and reliability, but also holds promise for scalable deployment in more complex and practical real-world scenarios.

Our main contributions are:

\begin{itemize} 
\item We introduce a structured reasoning framework that transitions from implicit exploration to explicit process modeling through \textit{guideline extraction} and \textit{stepwise refinement}. %This design addresses three core challenges in complex reasoning—unstable reasoning paths, lack of error correction, and limited experiential reuse—by enabling global guidance, iterative correction, and learned generalization.

\item We propose and validate three mechanisms for structured reasoning: stepwise execution improves stability, refinement enables error correction, and experience-based learning produces reusable strategies.

\item Our method consistently outperforms strong baselines on BBH tasks across model scales. We further investigate inter-model collaboration during refinement and demonstrate that our structured approach surpasses supervised fine-tuning (SFT) on LLaMA-3.3-70B, offering a scalable and interpretable alternative for complex reasoning.
\end{itemize}

\section{Related Work}
%\subsection{Automatic Prompt Optimization}
%Automatic prompt optimization (AutoPrompt) methods aim to refine input prompts dynamically, improving model adaptability in in-context learning (ICL). APE optimizes prompts through candidate generation, selection, and resampling. APO uses gradient-based refinement, while OPRO treats LLMs as feedback optimizers for iterative prompt improvement. PROMST incorporates human feedback for selection. IPC enhances robustness by generating boundary case data and refining prompts based on feedback. LEAP integrates error feedback into structured guidelines, combining few-shot CoT to guide iterative reasoning.

%While these methods improve prompt efficiency, they focus on local refinements rather than structuring the entire reasoning process. Our approach introduces stepwise guidance, ensuring more stable and interpretable adaptation.
\subsection{In-Context Learning}
In-Context Learning (ICL)~\cite{llmsurvey,reason_llm_survey} allows LLMs to perform tasks by conditioning on input examples without updating parameters~\cite{zero-shotcot,few-shot_icl,icl_survey,booststep}. Auto-CoT~\cite{auto-cot} enriches prompts by generating reasoning chains, while Many-shot~\cite{manyshot} and From Few to Many~\cite{fromfewtomany} improve performance by selecting better exemplars, which often via optimization over example pools.
Despite better exemplars, prior ICL methods fail to abstract reasoning structure—our approach learns structured guidelines to guide stable and adaptive reasoning.

\subsection{Chain-of-Thought Reasoning}
\iffalse
Chain-of-Thought (CoT)~\cite{cot,reason_llm_survey,hao2023reasoning} breaks down complex problems into step-wise logic, improving interpretability. ReAct~\cite{react} combines reasoning with actions for interactive problem solving. Tree-of-Thought (ToT) and its variants~\cite{tot,got,fot} organize reasoning as tree or graph search, enabling multi-path exploration and refinement.

Recent methods incorporate Monte Carlo Tree Search (MCTS)~\cite{mcts} to guide search. Beats~\cite{beats} uses MCTS with majority voting to enhance diversity; FoT~\cite{fot} extends this to a multi-tree setting; HiAR-ICL~\cite{hiar-icl} retrieves reusable patterns to maintain consistency.
While such methods expand the exploration space, they often overlook step-level correction. In contrast, our method extracts structured plans and applies step-wise refinement, enabling more stable and efficient reasoning.
\fi 

Chain-of-Thought (CoT)~\cite{cot,reason_llm_survey,hao2023reasoning} enhances interpretability via step-wise reasoning, while ReAct~\cite{react} integrates reasoning with actions. Tree-of-Thought (ToT) and its variants~\cite{tot,got,fot} structure inference as multi-path search.
Recent methods like Beats~\cite{beats}, FoT~\cite{fot}, and HiAR-ICL~\cite{hiar-icl} employ MCTS to improve exploration and consistency. However, they often overlook fine-grained step correction. In contrast, our method extracts structured plans and applies step-wise refinement for more stable and efficient reasoning.

\section{Method}

We propose a structured reasoning framework that transitions from implicit exploration to explicit, guideline-driven reasoning. Our method is built on two complementary components:

\textbf{Guideline Learning}: Automatically extracts stable reasoning patterns from correct examples and summarizes reflective signals from failure cases to form structured, step-wise guidelines.

\textbf{Guided Execution with Refinement}: Leverages learned guidelines to conduct reasoning step-by-step during inference, with refinement applied to each intermediate result for enhanced robustness.

\begin{figure*}[h]
    \centering
    \includegraphics[width=\textwidth]{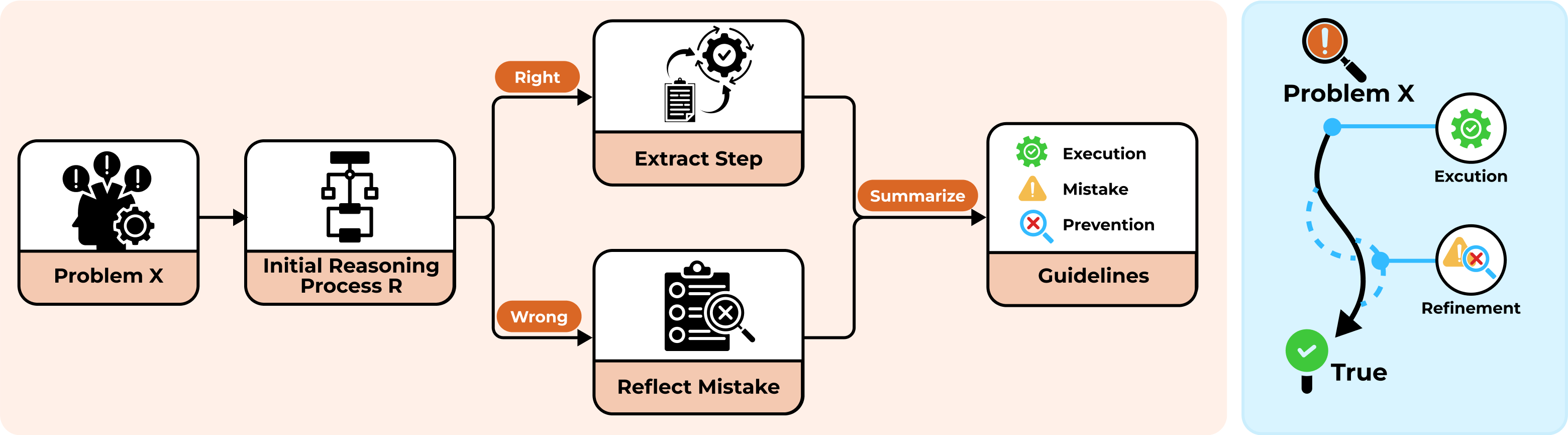}
    \caption{Overview of our framework. \textbf{Left:} The Guideline Learning module extracts reasoning steps from successful cases and identifies common mistakes from failed ones, summarizing them into generalizable \textit{guidelines}. \textbf{Right:} During inference, the model follows the learned guidelines step by step, with continuous \textit{refinement} to correct errors and improve reasoning stability.}
    \label{fig:framework}
\end{figure*}

\subsection{Guideline Learning}

To improve reasoning reliability, we introduce an automatic guideline learning module that distills structured reasoning steps from both correct and incorrect model trajectories.

Given an input $x$, the model first generates an initial reasoning process $R = (r_1, r_2, \dots, r_T,\hat{y})$, where $r_t$ corresponds to the  $t$-th step in the reasoning process. The final output $\hat{y}$ is then compared with the ground-truth label $y^*$:

- \textbf{If $\hat{y} = y^*$}, we extract key reasoning steps and summarize stable reasoning patterns.

- \textbf{If $\hat{y} \neq y^*$}, we analyze the trajectory to identify common reasoning failures and potential improvements.

We define two functions:
$f_{\text{ext}}$ extracts effective reasoning patterns from $(x, R, y^*)$, producing guideline candidates $\mathcal{G}$. $f_{\text{ref}}$ analyzes failures in $(x, R, y^*)$ to generate mistake-aware reflections $\mathcal{M}$:
\[
\mathcal{G} = f_{\text{ext}}(x,R, y^*)
\]
\[
\mathcal{M} = f_{\text{ref}}(x,R, y^*)
\]

After processing all samples in the train-set of dataset $\mathcal{D}$, we aggregate the learned guidelines into a final structured guideline steps $\mathcal{G}_T$,where $f_{\text{agg}}$ induces patterns across all examples to form $\mathcal{G}_T=(\mathcal{G}_1, \mathcal{G}_2, \dots, \mathcal{G}_T)$, among them $T$ is the length of reasoning steps.
\[
\mathcal{G}_T = f_{\text{agg}}(\mathcal{G}^{(1)}, \mathcal{G}^{(2)}, \dots, \mathcal{G}^{(N)})
\]

Here, $\mathcal{G}_T = (\mathcal{G}_1, \mathcal{G}_2, \dots, \mathcal{G}_T)$ represents a stepwise reasoning guideline synthesized across all examples.

\begin{algorithm}
\caption{Guideline Learning}
\label{alg:guideline}
\begin{algorithmic}[1]
\Require Training set $\mathcal{D}$ with samples $(x, y^*)$
\State Initialize reasoning guideline buffer 
\State $\mathcal{G}_{\text{buffer}} \gets \emptyset$
\For{each sample $(x, y^*) \in \mathcal{D}$}
    \State Generate initial reasoning path 
    \State $R = (r_1, r_2, \dots, r_T,\hat{y})$
    \If{$\hat{y} = y^*$} 
        \State Extract reasoning steps
        \State$\mathcal{G} = f_{\text{ext}}(x,R, y^*)$
    \Else
        \State Reflect on errors and prevent mistakes
        \State$\mathcal{M} = f_{\text{ref}}(x,R, y^*)$
    \EndIf
    \State Store: 
    \State$\mathcal{G}_{\text{buffer}} \gets \mathcal{G}_{\text{buffer}} \cup \mathcal{G}  \ or\  \mathcal{M}$
\EndFor
\State Aggregate final guideline: $\mathcal{G}_T = f_{\text{agg}}(\mathcal{G}_{\text{buffer}})$
\State \Return Learned guideline steps $\mathcal{G}_T$ 
\end{algorithmic}
\end{algorithm}

\subsection{Guided Execution with Refinement}

Once guidelines $\mathcal{G}_T$ are learned, we use them to guide step-wise inference and perform dynamic refinements to ensure reasoning stability.

For an input $x$ and learned guideline set $\mathcal{G}_T$, the model executes reasoning step-by-step:
\[
r_t = f_{\text{execute}}(x, \mathcal{G}_t)
\]
where $f_{\text{execute}}$ generates the current reasoning step $r_t$ based on input $x$ and the structured guideline $\mathcal{G}_T$.

After executing each step $r_t$, we inspect the result for common errors captured in the guideline $\mathcal{G}_t$. When an issue is found, we apply the associated prevention strategy to produce a refined step.

We define an function $f_{\text{refine}}$ that refines the reasoning step:
\[
r_t^* = f_{\text{refine}}(x,r_t, \mathcal{G}_t)
\]
where $r_t^*$ is the refined step that improves reasoning stability.

After all reasoning steps are executed and refined, the final output is derived as,where $R
=[r_1,r_2,...,r_t]$ is the final refined reasoning steps:
\[
\hat{y} = f_{\text{final}}(R)
\]
where $f_{\text{final}}$ integrates the reasoning sequence $R$ to extract the final answer $\hat{y} $.

\begin{algorithm}
\caption{Guided Execution with Refinement}
\label{alg:reasoning}
\begin{algorithmic}[1]
\Require Input $x$, learned guideline steps $\mathcal{G}_T$
\State Initialize reasoning path $R \gets \emptyset$
\For{each step $t$}
    \State Execute current step
    \State $r_t = \text{execute}(x, R,\mathcal{G}_t)$
    \State Append to reasoning path
    \State $R \gets R \cup \{r_t\}$
    \State Refine step result: 
    \State$r_t \gets f_{\text{refine}}(x,r_t, \mathcal{G}_t)$
\EndFor
\State Extract final conclusion $y = f_{\text{final}}(R)$
\State \Return final output $y$
\end{algorithmic}
\end{algorithm}

\section{Experiments}

\iffalse
We conduct experiments to evaluate the effectiveness of our proposed framework. We first compare our method with standard Chain-of-Thought (CoT) prompting and advanced reasoning frameworks including ReAct, Tree-of-Thought (ToT), and Beats (MCTS-based), which represent increasingly structured approaches to multi-step reasoning.

We then perform detailed analysis of three core mechanisms in our method. Step-wise execution improves path stability through explicit guidance, refinement corrects errors as they arise, and learning from prior reasoning enables models to generalize effective strategies beyond implicit trial-and-error.

Finally, we explore the extensibility of our framework through two settings: (1) a distillation scenario, where guidelines extracted from larger models are reused by smaller models, and (2) a model combination setup, where reasoning execution and refinement are handled by different models. These studies demonstrate the modularity and coordination potential of our approach.
\fi

\subsection{Experimental Setup}
\paragraph{Datasets}  
We primarily conduct experiments on eight diverse subsets from the BBH benchmark: Causal Judgement, Formal Fallacies, Geometric Shapes, Hyperbaton, Logical Deduction\_7, Navigate, Salient Translation Error Detection, and Multi-step Arithmetic. For clarity, we group these tasks into three categories and use their abbreviations throughout the paper. Following a consistent protocol, we randomly select 25\% of each dataset for training (guideline extraction) and use the remaining 75\% for evaluation, with all reported results based on the held-out test sets.

\textbf{Mathematical Reasoning:} Geometric Shapes (GS), Multi-step Arithmetic (MA), Navigate (NA)

\textbf{Logical Reasoning:} Causal Judgement (CJ), Formal Fallacies (FF), Logical Deduction\_7 (LD)

\textbf{Content Understanding:} Hyperbaton (HY), Salient Translation Error Detection (ST)

To further assess generality beyond BBH, we additionally evaluate on GSM8K, MATH-500, MBPP, and HumanEval. For all these datasets, we follow the same sampling strategy: 25\% of the data is used for training guideline extraction, and the remaining 75\% for evaluation.

\paragraph{Models}
We evaluate on both proprietary and open-source language models. GPT-4o and GPT-4o-mini serve as our primary closed-source models, while LLaMA-3.1-8B-Instruct and Qwen3-8B represent strong open-weight baselines. To study the transferability of structured reasoning, especially in distilling long CoT chains, we also include LLaMA-3.3-70B-Instruct and DeepSeek-R1-Distill-LLaMA-70B. For brevity, we refer to GPT-4o, GPT-4o-mini, LLaMA-3.1-8B-Instruct, and Qwen3-8B as 4o, mini, llama, and qwen, respectively.

\paragraph{Baselines}We compare against two categories of baselines:

\textbf{CoT-based Methods.} CoT~\cite{cot} prompts models to solve problems step by step. Few-shot CoT further provides annotated examples to guide human-like reasoning through demonstration.

\textbf{Reasoning Frameworks.} ReAct~\cite{react} combines reasoning with actions and observations for interactive problem solving. Tree-of-Thought (ToT)~\cite{tot} explores multiple solution paths via tree search ($n=3$). Beats~\cite{beats} uses MCTS with majority voting to enhance diversity and robustness ($d_{max}=5$, $a_{max}=3$). FoT~\cite{fot} adopts MCTS within a multi-tree reasoning structure ($n=3$, $iter_{max}=2$).

\paragraph{Evaluation Metrics}  
All models are evaluated using \textbf{accuracy}, where the model-generated answer is extracted from the \texttt{<answer>} tag. This ensures consistent answer alignment and allows for an objective comparison across different models and reasoning approaches.

\subsection{Comparison with Chain-of-Thought}
\begin{figure*}[tb]
	\centering
	\includegraphics[scale=0.6]{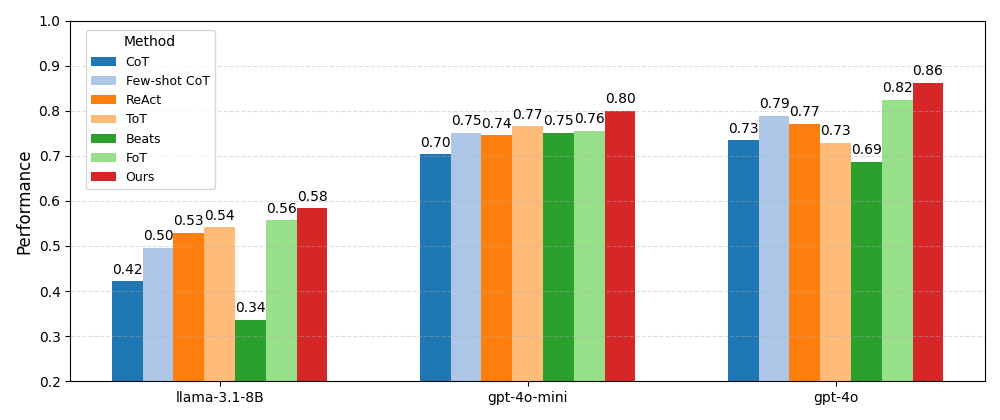}
          \caption{
          \textbf{Overall performance comparison across six reasoning methods under different model scales}. The methods include: \textbf{CoT-based} approaches (CoT, Few-shot CoT), which rely on implicit pattern imitation; and \textbf{reasoning frameworks} (ReAct, ToT, Beats,FoT), which introduce dynamic interaction, search, or voting-based selection. Our structured approach (Ours) consistently achieves superior performance, demonstrating the effectiveness of guideline-based execution and stepwise refinement in complex reasoning tasks.
          }
	\label{fig:comp}
\end{figure*}

As illustrated in Table~\ref{tab:cot_comparison},Table~\ref{tab:extra_results} and Figure~\ref{fig:comp}, our method consistently surpasses both CoT and Few-shot CoT across eight reasoning tasks, spanning mathematical, logical, and content understanding categories. The gains are especially pronounced on multi-step and long-horizon problems, and remain robust across model scales from LLaMA-3.1-8B to GPT-4o.

These findings highlight the limitations of conventional CoT approaches, which rely on implicit reasoning and static demonstrations. Even with few-shot examples, models often fail to derive transferable strategies or to recover from accumulated errors, resulting in unstable reasoning. In contrast, our framework integrates structured guidelines to direct each step and a refinement mechanism for adaptive correction, producing more stable and accurate performance.

\begin{table*}[h]
    \centering
    \footnotesize
    \caption{Performance comparison between CoT-based methods and our approach across eight reasoning tasks. The datasets are grouped into three categories: Mathematical Reasoning (GS, MA, NA), Logical Reasoning (CJ, FF, LD), and Content Understanding (HY, ST).}
    \label{tab:cot_comparison}
    \renewcommand{\arraystretch}{1.2}
    \setlength{\tabcolsep}{5pt}  % 可微调列间距
    
    \resizebox{\textwidth}{!}{
    \begin{tabular}{ll*{9}{c}}
        \toprule
        \textbf{Model} & \textbf{Method} & \multicolumn{3}{c}{\textbf{Mathematical Reasoning}} & \multicolumn{3}{c}{\textbf{Logical Reasoning}} & \multicolumn{2}{c}{\textbf{Content Understanding}} & \textbf{Avg} \\
        \cmidrule(lr){3-5} \cmidrule(lr){6-8} \cmidrule(lr){9-10}
        & & GS & MA & NA & CJ & FF & LD & HY & ST & \\
        \midrule
        \multirow{3}{*}{4o} 
        & CoT & 0.631 & 0.968 & 0.979 & 0.621 & 0.476 & 0.620 & 0.882 & 0.695 & 0.734 \\
        & Few-shot CoT & 0.599 & 0.978 & 0.973 & 0.593 & 0.636 & 0.813 & 0.989 & 0.787 & 0.789 \\
        & \textbf{Ours} & 0.711 & 0.963 & 0.973 & 0.671 & 0.770 & 0.925 & 0.995 & 0.888 & \textbf{0.862} \\
        \midrule
        \multirow{3}{*}{mini} 
        & CoT & 0.455 & 0.957 & 0.968 & 0.636 & 0.743 & 0.508 & 0.775 & 0.583 & 0.703 \\
        & Few-shot CoT & 0.551 & 0.930 & 0.963 & 0.636 & 0.733 & 0.604 & 0.957 & 0.631 & 0.751 \\
        & \textbf{Ours} & 0.658 & 0.936 & 0.952 & 0.657 & 0.679 & 0.711 & 0.995 & 0.813 & \textbf{0.800} \\
        \midrule
        \multirow{3}{*}{llama} 
        & CoT & 0.267 & 0.428 & 0.497 & 0.493 & 0.251 & 0.272 & 0.529 & 0.639 & 0.422 \\
        & Few-shot CoT & 0.679 & 0.455 & 0.674 & 0.507 & 0.369 & 0.203 & 0.513 & 0.567 & 0.496 \\
        & \textbf{Ours} & 0.583 & 0.401 & 0.529 & 0.621 & 0.444 & 0.572 & 0.807 & 0.717 & \textbf{0.584} \\
        \bottomrule
    \end{tabular}
    }
\end{table*}

\subsection{Comparison with Reasoning Frameworks}
As shown in Table~\ref{tab:comp_reason} and Figure~\ref{fig:comp}, our method consistently outperforms ReAct, Tree-of-Thought (ToT), Beats, and FoT across three model scales: LLaMA-3.1-8B, GPT-4o-mini, and GPT-4o. The gains hold across all BBH task categories—mathematical, logical, and content understanding—demonstrating the robustness of structured guideline-based reasoning. Furthermore, Table~\ref{tab:extra_results} shows that these improvements generalize beyond BBH, with the mini model achieving strong results on mathematical (GSM8K, MATH-500) and code generation (MBPP, HumanEval) benchmarks.

These results suggest that guideline-driven reasoning provides a more stable and efficient alternative to existing frameworks. While ReAct and ToT rely on trial-and-error exploration or heuristic tree search, and Beats on costly MCTS sampling, FoT improves performance through multi-tree exploration with limited reuse of thought processes. In contrast, our framework unifies global planning via learned guidelines with local adaptability through refinement, enabling targeted optimization and dynamic correction—particularly effective for multi-step reasoning tasks.

\begin{table*}[htb]
    \centering
    \footnotesize
    \caption{Performance comparison between our approach and other reasoning frameworks across eight reasoning tasks, grouped by category.}
    \label{tab:comp_reason}
    \renewcommand{\arraystretch}{1.2}
    \setlength{\tabcolsep}{5pt}

    \resizebox{\textwidth}{!}{
    \begin{tabular}{llccccccccc}
        \toprule
        \textbf{Model} & \textbf{Method} & \multicolumn{3}{c}{\textbf{Mathematical Reasoning}} & \multicolumn{3}{c}{\textbf{Logical Reasoning}} & \multicolumn{2}{c}{\textbf{Content Understanding}} & \textbf{Avg} \\
        \cmidrule(lr){3-5} \cmidrule(lr){6-8} \cmidrule(lr){9-10}
        & & GS & MA & NA & CJ & FF & LD & HY & ST & \\
        \midrule
        \multirow{4}{*}{4o} 
        & ReAct & 0.545 & 0.620 & 0.995 & 0.619 & 0.807 & 0.925 & 0.872 & 0.786 & 0.771 \\
        & ToT & 0.419 & 0.663 & 0.968 & 0.669 & 0.775 & 0.882 & 0.674 & 0.786 & 0.729 \\
        & Beats & 0.545 & 0.989 & 0.984 & 0.633 & 0.283 & 0.684 & 0.850 & 0.524 & 0.687 \\
        & FoT & 0.738 &0.973 &0.984&0.679&0.695&0.834&0.947&0.743&0.824 \\
        & \textbf{Ours} & 0.711 & 0.963 & 0.973 & 0.671 & 0.770 & 0.925 & 0.995 & 0.888 & \textbf{0.862} \\
        \midrule
        \multirow{4}{*}{mini} 
        & ReAct & 0.481 & 0.952 & 0.952 & 0.669 & 0.690 & 0.802 & 0.775 & 0.636 & 0.745 \\
        & ToT & 0.561 & 0.941 & 0.979 & 0.657 & 0.738 & 0.706 & 0.882 & 0.663 & 0.766 \\
        & Beats & 0.545 & 0.947 & 0.979 & 0.679 & 0.551 & 0.727 & 0.963 & 0.620 & 0.751 \\
        & FoT & 0.486& 0.947 &0.979&0.640&0.775&0.733 &0.845&0.636 &0.755 \\
        & \textbf{Ours} & 0.658 & 0.936 & 0.952 & 0.657 & 0.679 & 0.711 & 0.995 & 0.813 & \textbf{0.800} \\
        \midrule
        \multirow{4}{*}{llama} 
        & ReAct & 0.417 & 0.465 & 0.454 & 0.471 & 0.524 & 0.460 & 0.780 & 0.663 & 0.529 \\
        & ToT & 0.476 & 0.439 & 0.561 & 0.450 & 0.588 & 0.449 & 0.733 & 0.642 & 0.542 \\
        & Beats & 0.209 & 0.267 & 0.652 & 0.471 & 0.241 & 0.241 & 0.310 & 0.299 & 0.336 \\
        & FoT & 0.380 & 0.444& 0.620 &0.593 &0.545& 0.529 & 0.706 &0.636 &0.557 \\
        & \textbf{Ours} & 0.583 & 0.401 & 0.529 & 0.621 & 0.444 & 0.572 & 0.807 & 0.717 & \textbf{0.584} \\
        \bottomrule
    \end{tabular}
    }
\end{table*}

\begin{table}[h]
\centering
\caption{Performance on math (GSM8K, MATH-500) and code (MBPP, HumanEval) benchmarks using the \textbf{mini} model.}
\label{tab:extra_results}
\small
\begin{tabular}{lcc|cc}
\toprule
\multirow{2}{*}{\textbf{Method}} & \multicolumn{2}{c|}{\textbf{Math}} & \multicolumn{2}{c}{\textbf{Code}} \\
\cmidrule(lr){2-3} \cmidrule(lr){4-5}
 & GSM8K & MATH-500 & MBPP & HumanEval \\
\midrule
CoT   & 0.872 & 0.578 & 0.663 & 0.886 \\
ReAct & 0.782 & 0.580 & 0.690 & 0.920 \\
ToT   & 0.901 & 0.652 & 0.673 & 0.900 \\
Beats & 0.906 & 0.659 & 0.700 & 0.935 \\
FoT   & 0.904 & 0.658 & 0.691 & 0.931 \\
Ours  & \textbf{0.921} & \textbf{0.676} & \textbf{0.730} & \textbf{0.938} \\
\bottomrule
\end{tabular}
\normalsize
\end{table}

\subsection{Detailed Analysis}

We conduct a comprehensive analysis of the three core mechanisms in our framework: stepwise execution, refinement, and guideline-based learning. These mechanisms are evaluated for their impact on performance, stability, and generalizability in reasoning tasks. We present ablation studies and  inter-collaboration patterns under different model configurations.

Table~\ref{tab:ablation_summary} summarizes the performance under different configurations. Each setting toggles whether guideline is learned from past experience, stepwise execution, and refinement are enabled. Results are reported for GPT-4o, GPT-4o-Mini, LLaMA-3.1-8B and Qwen3-8B.

\begin{table}[h]
\centering
\caption{Performance comparison across different configurations. \textcolor{green}{\ding{51}} indicates the component is enabled, \textcolor{red}{\ding{55}} means disabled.}
\label{tab:ablation_summary}
\small
\begin{tabular}{ccccccc}
\toprule
\textbf{Learn} & \textbf{Step} & \textbf{Refine} & \textbf{4o} & \textbf{mini} & \textbf{llama} & \textbf{qwen} \\
\midrule
\textcolor{green}{\ding{51}} & \textcolor{green}{\ding{51}} & \textcolor{green}{\ding{51}} & \textbf{0.862} & \textbf{0.800} & 0.584 & \textbf{0.820} \\
\textcolor{green}{\ding{51}} & \textcolor{green}{\ding{51}} & \textcolor{red}{\ding{55}} & 0.802 & 0.785 & 0.635 & 0.803 \\
\textcolor{green}{\ding{51}} & \textcolor{red}{\ding{55}} & \textcolor{green}{\ding{51}} & 0.801 & 0.762 & 0.646 & 0.787 \\
\textcolor{green}{\ding{51}} & \textcolor{red}{\ding{55}} & \textcolor{red}{\ding{55}} & 0.799 & 0.760 & \textbf{0.650} & 0.770 \\
\textcolor{red}{\ding{55}} & \textcolor{green}{\ding{51}} & \textcolor{green}{\ding{51}} & 0.809 & 0.634 & 0.499 & -- \\
\bottomrule
\end{tabular}
\end{table}

\subsubsection{Stepwise Execution Mechanism}
According to Table~\ref{tab:ablation_summary}, for both the 4o and mini models, stepwise execution consistently outperforms single-step reasoning, regardless of whether the refinement mechanism is enabled. This highlights the benefit of structuring the reasoning process into sequential steps to enhance stability and robustness.

However, an interesting deviation is observed with the llama model, which performs worse under full refinement compared to simpler settings. In contrast, the similarly sized qwen model does not exhibit this degradation, suggesting that the effect is not purely due to model scale but rather tied to differences in reasoning capability. We further investigate this phenomenon in Section~\ref{inter-model} through inter-model execution and refinement experiments.

\begin{figure}[h]
  \centering
  \includegraphics[width=0.9\linewidth]{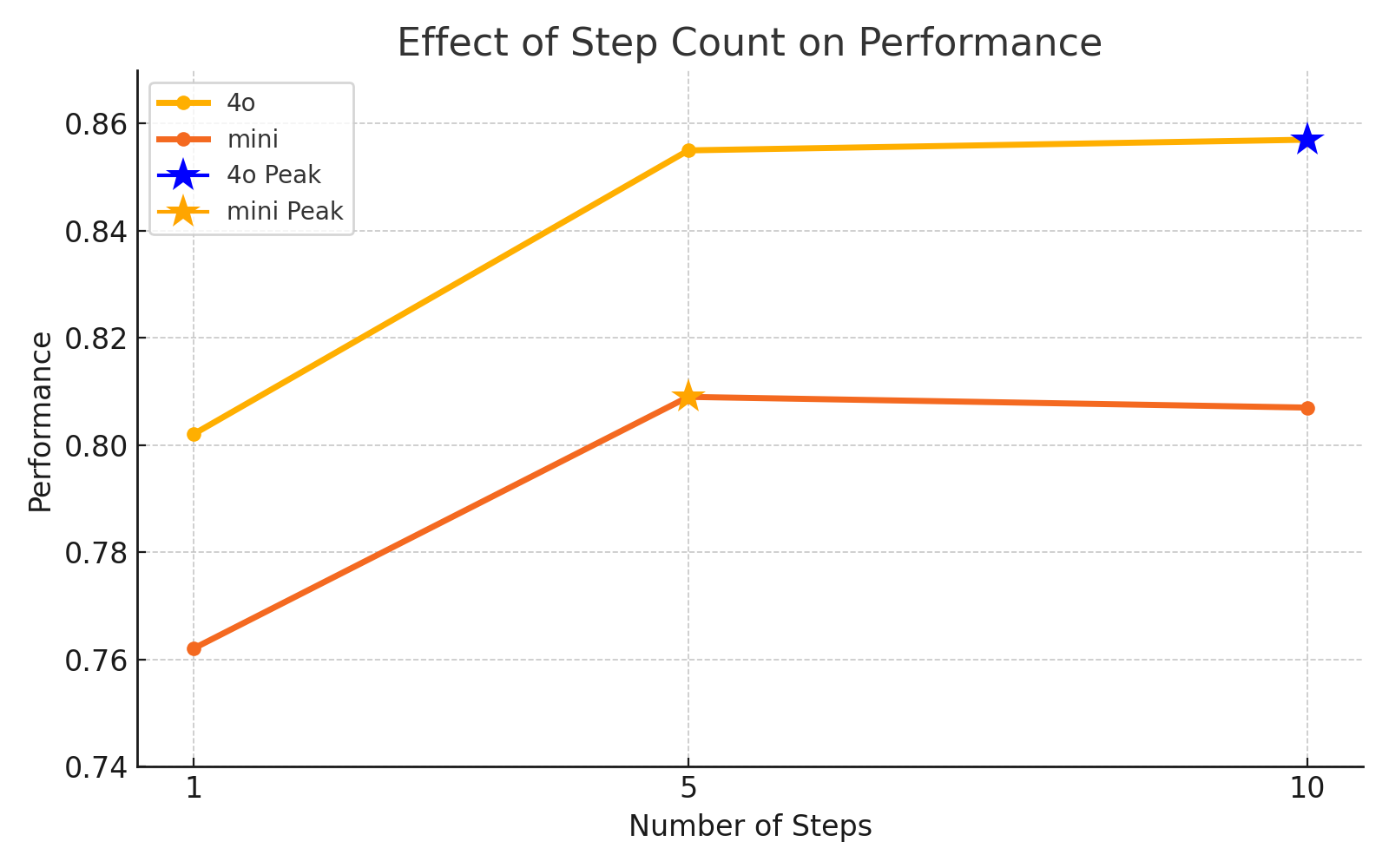}
  \caption{Effect of step count on performance}
  \label{fig:step_cmp}
\end{figure}

To better understand the role of execution granularity, we examine the impact of step count using both the 4o and mini models (note: in Table~\ref{tab:ablation_summary}, the number of steps is not strictly fixed but varies by task, typically ranging from 6 to 10). As shown in Figure~\ref{fig:step_cmp}, increasing the number of reasoning steps from one to five yields substantial performance gains, suggesting that stepwise decomposition enhances stability and decision quality. Beyond five steps, however, the improvement plateaus, indicating diminishing returns.
%\begin{table}[h]
%\centering
%\caption{Effect of step count on performance.}
%\label{tab:step_num_table}
%\begin{tabular}{lccc}
%\toprule
%\textbf{Model} & \textbf{1 step} & \textbf{5 steps} & \textbf{10 steps} \\
%\midrule
%4o & 0.802 & 0.855 & \textbf{0.857} \\
%mini & 0.762 & \textbf{0.809} & 0.807 \\
%\bottomrule
%\end{tabular}
%\end{table}

\subsubsection{Refinement Mechanism}
As presented in Table~\ref{tab:ablation_summary}, applying refinement improves performance across both stepwise and single-step execution, indicating its general effectiveness in enhancing inference quality. While this pattern holds for 4o, Mini, and qwen, LLaMA exhibits a slight performance drop when refinement is applied, suggesting that the issue is not simply due to model size but rather reflects differences in reasoning robustness.

\begin{figure}[h]
  \centering
  \includegraphics[width=1.0\linewidth]{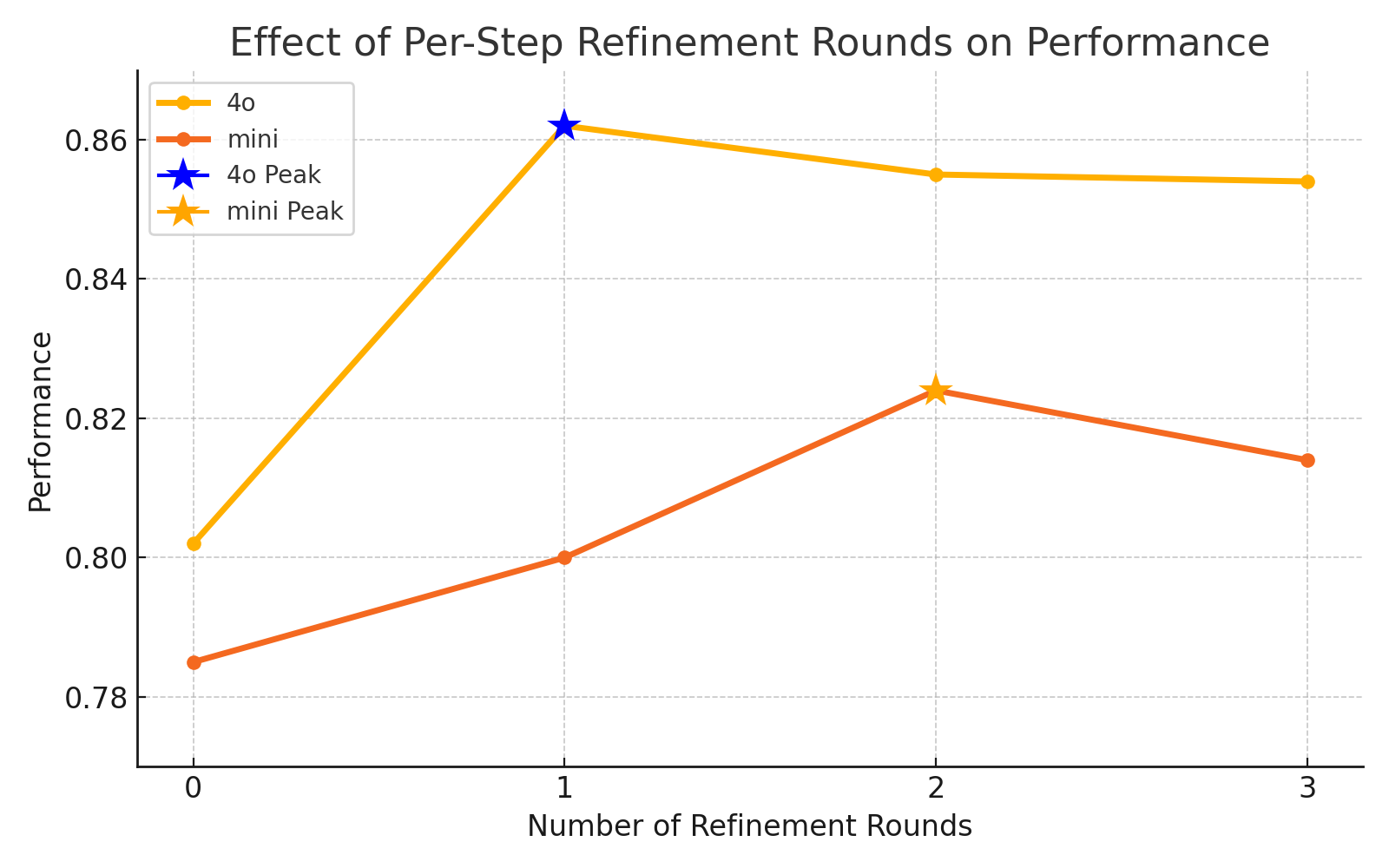}
  \caption{Effect of Per-Step Refinement Rounds on Performance}
  \label{fig:refine_round}
\end{figure}

To further examine this effect, we vary the number of refinement rounds, as shown in Figure~\ref{fig:refine_round}. For 4o, performance peaks after a single refinement, implying that one iteration of self-correction is sufficient for strong models to stabilize their reasoning. In contrast, Mini achieves optimal performance with two refinement rounds, indicating that moderately sized models benefit from additional correction to offset weaker initial outputs. However, further refinement beyond the optimal point introduces noise or over-adjustment, leading to performance degradation in both models. These findings highlight the importance of calibrating refinement depth based on model capacity.

%\begin{table}[h]
%\centering
%\caption{Effect of Per-Step Refinement Rounds on Performance}
%\label{tab:refine_num}
%\begin{tabular}{lcccc}
%\toprule
%\textbf{Model} & \textbf{0} & \textbf{1} & \textbf{2} & \textbf{3} \\
%\midrule
%GPT-4o & 0.802 & 0.862 & 0.855 & 0.854 \\
%Mini & 0.785 & 0.800 & 0.824 & 0.814 \\
%\bottomrule
%\end{tabular}
%\end{table}

\subsubsection{Inter-Model Collaboration}
\label{inter-model}
Table~\ref{tab:ablation_summary} shows both 4o and mini benefit from the addition of stepwise execution and refinement mechanisms, exhibiting consistent performance improvements. In contrast, the performance of llama degrades when either mechanism is introduced. This suggests that smaller models may suffer from error accumulation and lack sufficient capacity for effective refinement.

\begin{figure}[h]
\centering
\includegraphics[width=1.0\linewidth]{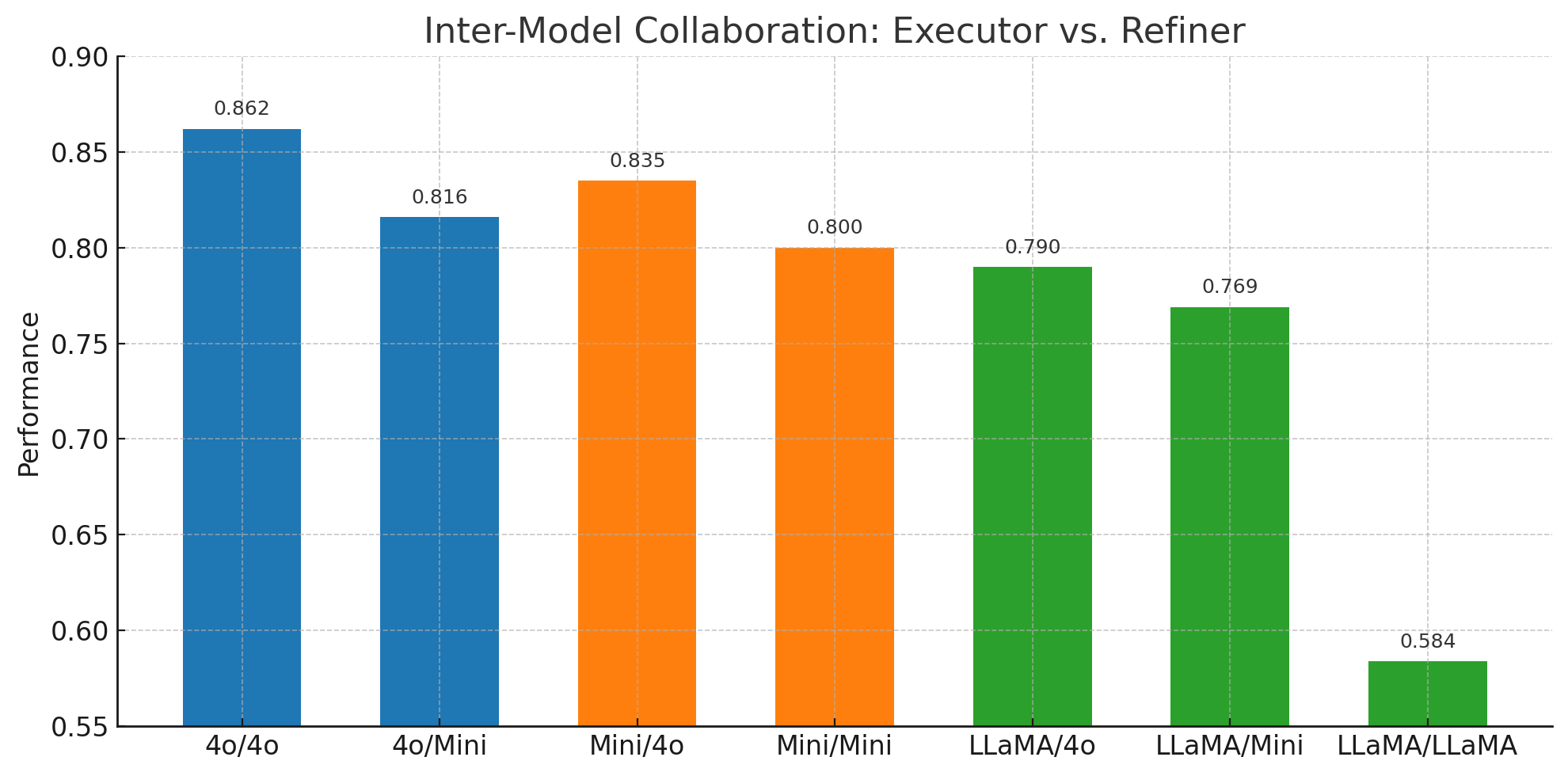}
\caption{Performance of inter-model collaboration under different executor/refiner pairings. Models are grouped by the executor (4o, mini, llama). Within each group, using a stronger refiner generally improves performance.}
\label{fig:inter_model}
\end{figure}

To overcome this limitation, we explore \textit{inter-model collaboration} by pairing models as executor and refiner. As shown in Figure~\ref{fig:inter_model}, stronger refiners like GPT-4o consistently improve performance, while weaker refiners (e.g., LLaMA or mini) may degrade results, underscoring the need for complementary model roles.

\iffalse
\begin{table}[h]
\centering
\caption{Performance of Inter-Model Collaboration: Reasoning is executed with one model and refined by another.}
\label{tab:inter_model}
\begin{tabular}{lcccccc}
\toprule
\textbf{Executor/Refiner} & \textbf{4o/4o} & \textbf{4o/Mini} & \textbf{Mini/Mini} & \textbf{Mini/4o} & \textbf{LLaMA/Mini} & \textbf{LLaMA/4o} \\
\midrule
\textbf{Performance} & \textbf{0.862} & 0.816 & 0.800 & 0.835 & 0.769 & 0.790 \\
\bottomrule
\end{tabular}
\end{table}
\fi

\subsubsection{Supervision and Transferability}

A key question is whether our method requires perfectly task-aligned supervision. We evaluate guideline transferability in two settings:  

(A) \emph{In-domain transfer}: each test task uses guidelines from another task in the same domain.  
-- MA (Math) from GS  
-- CJ (Logic) from FF  
-- HY (Content) from ST  

(B) \emph{Cross-domain transfer}: guidelines are transferred across domains, with GS (Math), FF (Logic), and ST (Content) serving as representative sources.  

Results are shown in Table~\ref{tab:guideline_transfer}.

\begin{table}[h]
\centering
\caption{Guideline transferability under (A) in-domain and (B) cross-domain settings (accuracy). Domain labels: Math = Mathematical, Logic = Logical, Content = Content Understanding.}
\label{tab:guideline_transfer}

{\bf (A) In-domain transfer}\vspace{3pt}

\begin{tabular}{lcc}
\toprule
\textbf{Task} & \textbf{Task-Specific} & \textbf{In-Domain} \\
\midrule
MA (Math)    & 0.936 & 0.957 \\
CJ (Logic)   & 0.657 & 0.620 \\
HY (Content) & 0.995 & 0.970 \\
\bottomrule
\end{tabular}

\vspace{6pt}
{\bf (B) Cross-domain transfer}\vspace{3pt}

\begin{tabular}{lccc}
\toprule
\textbf{Task} & \textbf{GS (Math)} & \textbf{FF (Logic)} & \textbf{ST (Content)} \\
\midrule
GS & 0.658 & 0.513 & 0.668 \\
FF & 0.599 & 0.679 & 0.535 \\
ST & 0.743 & 0.786 & 0.813 \\
\bottomrule
\end{tabular}
\end{table}

In-domain guidelines often perform on par with or better than task-specific ones, while cross-domain transfer also yields competitive results. These findings indicate that the method is robust to weaker supervision and moderate domain shifts.

\subsubsection{Learning vs. Self-Planning}

We compare models guided by learned guidelines with those that rely solely on implicit self-planning. In the self-plan setting, the model generates a step plan before execution, but lacks prior execution experience or task-specific guidance for avoiding common errors.

\begin{figure}[h]
\centering
\includegraphics[width=1.0\linewidth]{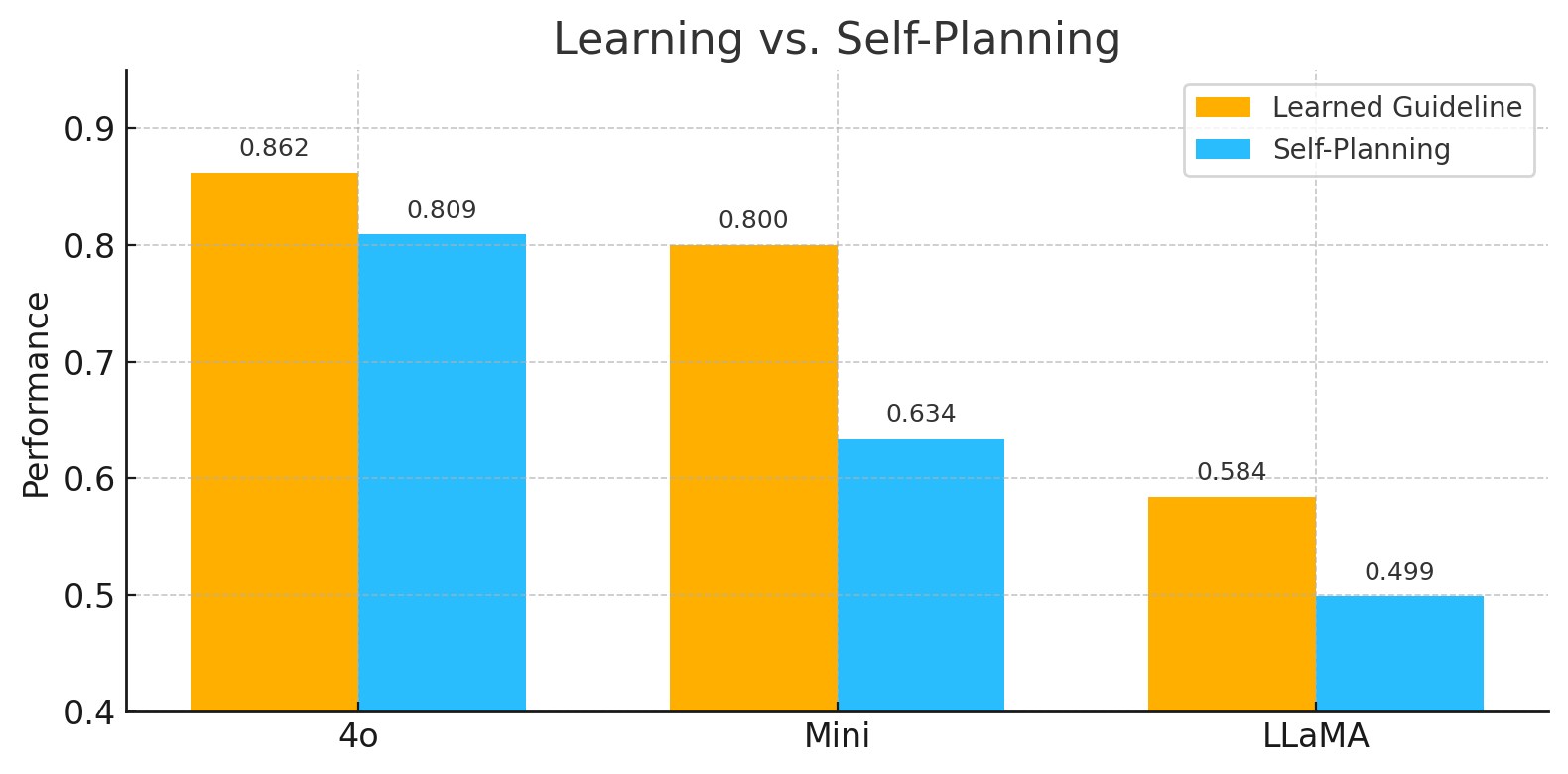}
\caption{Guideline-based reasoning consistently outperforms self-planning.}
\label{fig:learn_vs_plan}
\end{figure}

As depicted in the Figure~\ref{fig:learn_vs_plan}, structured reasoning driven by learned experience consistently outperforms such unguided planning. Even for large models like 4o, learning-based execution yields noticeable gains, while the gap becomes more pronounced for smaller models like mini and llama.

\subsubsection{SFT vs. Structured Reasoning}

Inspired by prior works~\cite{icl_optimize},we investigate whether structured reasoning can be effectively distilled from a stronger model and transferred to others. We compare two approaches: 	\textbf{(1) Guideline-Assisted Reasoning}, which extracts stepwise reasoning traces from R1's Chain-of-Thought (CoT) and applies them to other models; and 	\textbf{(2) SFT-Based Distillation}, which uses the distilled version of R1 (DeepSeek-R1-Distill-llama-3.3-70B) with standard CoT prompting to assess whether reasoning quality is preserved after fine-tuning.

\begin{figure}[h]
\centering
\includegraphics[width=1.0\linewidth]{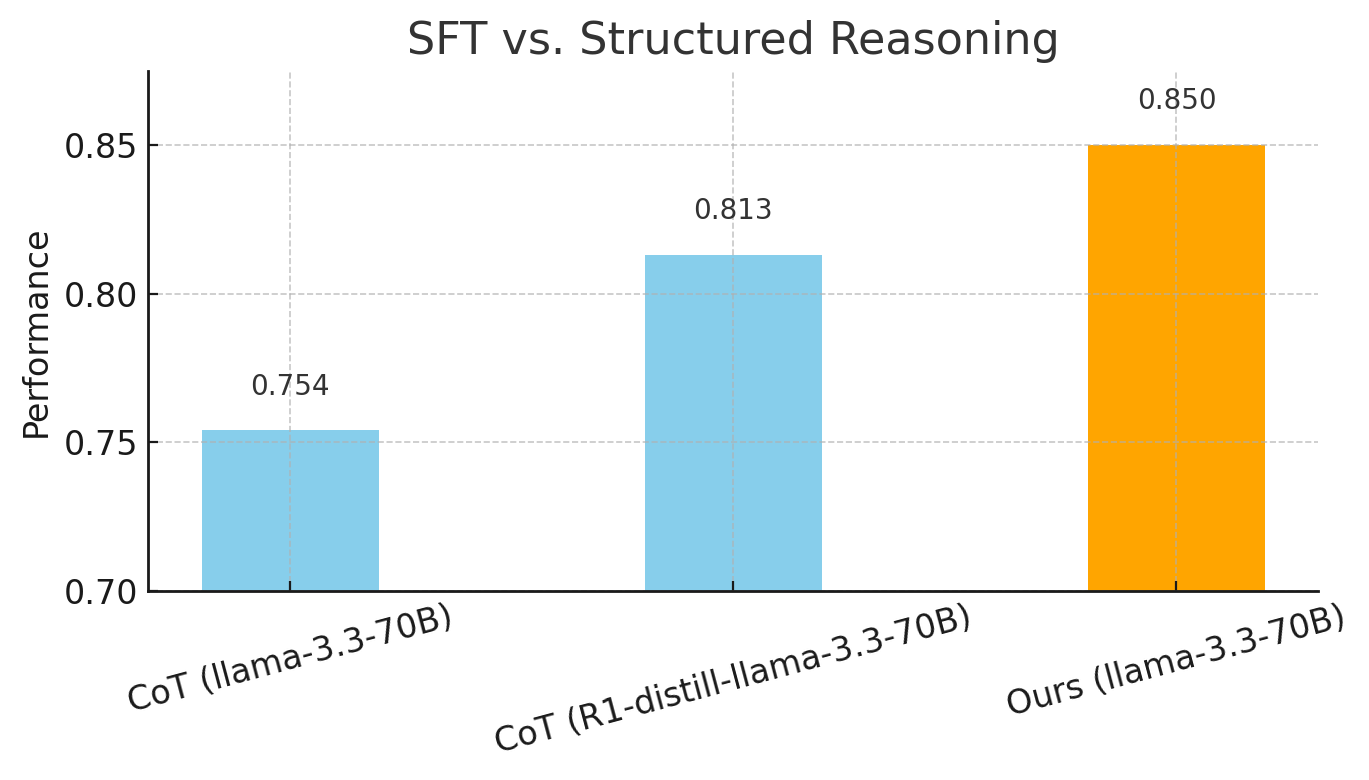}
\caption{\textbf{SFT  vs. Structured reasoning} Performance comparison between CoT prompting on the original model, its SFT-distilled variant, and our guideline-based reasoning framework (all using llama-3.3-70B).}
\label{fig:cmp_sft}
\end{figure}

As shown by Figure~\ref{fig:cmp_sft}, both approaches are evaluated on the same base model, llama-3.3-70B. Our guided reasoning framework consistently outperforms the SFT-distilled variant, despite requiring no additional training. This highlights that structured reasoning can be effectively induced through lightweight external guidance, providing a more interpretable and flexible alternative to implicit learning via supervised fine-tuning.

\iffalse
\begin{table*}[htb]
    \centering
    \footnotesize  % 缩小字体
    \caption{Performance comparison between our approach and React,ToT methods.}
    \label{tab:distillation}
    \renewcommand{\arraystretch}{1.2}
    \setlength{\tabcolsep}{4pt} % 调整列间距，仅影响数据列

    \resizebox{\textwidth}{!}{
    \begin{tabular}{l *{9}{p{1.5cm}}}  % 只让数据列均匀分布
        \toprule
        \textbf{Method} & \multicolumn{3}{c}{\textbf{Mathematical Reasoning}} & \multicolumn{3}{c}{\textbf{Logical Reasoning}} & \multicolumn{2}{c}{\textbf{Content Understanding}} & \textbf{Avg} \\
        \cmidrule(lr){2-4} \cmidrule(lr){5-7} \cmidrule(lr){8-9}
        & GS & MA & NA & CJ & FF & LD & HY & ST \\
        \midrule
        CoT (llama-3.3-70B) & 0.411 & 0.93 & 0.936 & 0.678 & 0.813 & 0.604 & 0.951 & 0.711 & 0.754 \\
        CoT (R1-distll-llama-3.3-70B) & 0.652 & 0.963 & 0.9/52 & 0.671 & 0.816 & 0.673 & 0.994    & 0.78 & 0.813\\
        Ours (lama-3.3-70B) & 0.69 & 0.963 & 0.98 & 0.7 & 0.828 & 0.91 & 0.979 & 0.749 & 0.85 \\
        \bottomrule
    \end{tabular}
    }
\end{table*}
\fi

\subsection{Case Study:Geometry Shapes}
To qualitatively evaluate the model's structured reasoning ability, we conduct a case study on the \texttt{geometric\_shapes} task. The guideline extracted from training examples is shown in Appendix Figure~\ref{fig:geo_guideline_final}, detailing step-wise instructions, common mistakes, and prevention strategies.  
Appendix Figure~\ref{fig:svg_step_refine_full} demonstrates how GPT-4o applies this guideline to a specific SVG input. Notably, an initial error in path closure detection is corrected through refinement, while all other steps are executed correctly.

\section{Conclusion}
We propose a structured reasoning framework that transitions from implicit exploration to explicit process modeling via guideline extraction and stepwise refinement. This approach enhances reasoning stability, supports error correction, and enables experience-based generalization. Experiments on eight BBH tasks across multiple models show consistent improvements over strong baselines, including CoT, ReAct, ToT, Beats and FoT. Further analysis highlights the contributions of stepwise execution, iterative refinement, and learned guidance. We also explore inter-model collaboration in the refinement stage and show that our method performs competitively with, and in some cases outperforms, supervised fine-tuning (SFT) approaches that distill knowledge from larger models—demonstrating strong scalability and practical potential for complex reasoning tasks.

\section*{Limitations}
Although we explore inter-model collaboration during refinement, the design space of combining models at different scales remains underexplored. Future work may investigate more diverse and adaptive model configurations to further enhance efficiency, robustness, and scalability in practical deployments.

\section*{Acknowledgments}

% \bibliography{anthology,references}
% \input{output.bbl}

\appendix
\newpage
\section{Appendix}
\newcommand{\correct}{\textcolor{green!60!black}{\ding{51}}}  % 对号
\newcommand{\wrong}{\textcolor{red!80!black}{\ding{55}}}      % 错号

\newcommand{\exec}{{\textbf{\textcolor{black}{Execution}}}}
\newcommand{\mistake}{{\textbf{\textcolor{red!60!black}{Mistake}}}}
\newcommand{\prevention}{{\textbf{\textcolor{teal!60!black}{Prevention}}}}

\subsection{Case Study: Structured Reasoning in Geometric Shapes Task}
\label{app:case-geo}

All results in this case study are generated using GPT-4o.  
Figure~\ref{fig:geo_guideline_final} outlines the structured guideline learned for reasoning over geometric shapes based on SVG path representations.  
Figure~\ref{fig:svg_step_refine_full} provides a step-wise case study where the model applies this guideline to identify a triangle, including correction of an intermediate reasoning error.

\begin{figure*}[ht]
    \centering
    \begin{tcolorbox}[
        width=0.95\textwidth,
        colframe=black!60,
        colback=gray!1,
        fonttitle=\bfseries,
        fontupper=\footnotesize,
        title={Structured Guideline from the \texttt{geometric\_shapes} Task},
        breakable,
        enhanced,
        sharp corners=southwest,
        boxrule=0.5pt,
        left=2mm, right=2mm, top=1mm, bottom=1mm,
    ]

    \textbf{Step 1: Parse and Map the SVG Path Commands} \\[-0.5em]
    \begin{tcolorbox}[colback=gray!3, colframe=gray!20, boxrule=0pt, left=1.5mm, right=1.5mm, top=0.5mm, bottom=0.5mm]
    \exec: Systematically parse all SVG path commands (e.g., \texttt{M}, \texttt{L}, \texttt{A}). Record vertices, edges, and arcs. Treat \texttt{M} commands as boundaries for disconnected sub-paths.\\[0.3em]
    \mistake: Misinterpreting \texttt{M} commands as continuous drawing instructions or including them as extra vertices.
    \prevention: Explicitly isolate sub-paths introduced by \texttt{M}. Visually trace commands to confirm disjointness and segment boundaries.
    \end{tcolorbox}

    \textbf{Step 2: Identify Path Closure and Count Vertices/Edges} \\[-0.5em]
    \begin{tcolorbox}[colback=gray!3, colframe=gray!20, boxrule=0pt, left=1.5mm, right=1.5mm, top=0.5mm, bottom=0.5mm]
    \exec: Check whether the final point matches the starting point. If closed, count distinct vertices and edges.\\[0.3em]
    \mistake: Overlooking closure or miscounting repeated vertices, resulting in incorrect edge totals.\\[0.3em]
    \prevention: Directly compare coordinates to verify closure. Use a deduplicated list of points to ensure accurate counting.
    \end{tcolorbox}

    \textbf{Step 3: Analyze Curved or Straight Line Features} \\[-0.5em]
    \begin{tcolorbox}[colback=gray!3, colframe=gray!20, boxrule=0pt, left=1.5mm, right=1.5mm, top=0.5mm, bottom=0.5mm]
    \exec: Distinguish straight lines (\texttt{L}) from arcs (\texttt{A}). Extract arc parameters such as radii, sweep flags, and angles.\\[0.3em]
    \mistake: Assuming all arcs are full circles or failing to recognize mixed shapes with curves and lines.\\[0.3em]
    \prevention: Analyze arc direction, symmetry parameters (\texttt{rx}, \texttt{ry}), and flags to differentiate ellipses, sectors, and polygons.
    \end{tcolorbox}

    \textbf{Step 4: Validate Shape Properties Against Geometric Definitions} \\[-0.5em]
    \begin{tcolorbox}[colback=gray!3, colframe=gray!20, boxrule=0pt, left=1.5mm, right=1.5mm, top=0.5mm, bottom=0.5mm]
    \exec: Match the figure to geometric definitions using vertex count, symmetry, angles, and side lengths (e.g., triangle, quadrilateral, ellipse).\\[0.3em]
    \mistake: Prematurely classifying a figure without checking supporting properties (e.g., mislabeling a quadrilateral as a kite).\\[0.3em]
    \prevention: Apply precise rules—verify symmetry, angle consistency, and equal-length sides where required.
    \end{tcolorbox}

    \textbf{Step 5: Apply a Process of Elimination to Confirm the Shape} \\[-0.5em]
    \begin{tcolorbox}[colback=gray!3, colframe=gray!20, boxrule=0pt, left=1.5mm, right=1.5mm, top=0.5mm, bottom=0.5mm]
    \exec: Compare observed features against candidate shapes and eliminate those that do not match.\\[0.3em]
    \mistake: Failing to rule out incompatible categories (e.g., misidentifying triangles as sectors due to curvature confusion).\\[0.3em]
    \prevention: Use structured elimination (e.g., decision trees) to ensure all remaining options match key shape attributes.
    \end{tcolorbox}

    \end{tcolorbox}
    \caption{A five-step structured guideline learned from the \texttt{geometric\_shapes} task. Each step includes execution instructions, common mistake patterns, and prevention strategies to improve SVG-based shape reasoning.}
    \label{fig:geo_guideline_final}
\end{figure*}

\begin{figure*}[ht]
\centering
\begin{tcolorbox}[
    width=0.95\textwidth,
    colframe=black!60,
    colback=gray!2,
    fonttitle=\bfseries,
    fontupper=\footnotesize,
    title={Structured Reasoning on SVG Path with Step-wise Refinement},
    breakable,
    enhanced,
    sharp corners=southwest,
    boxrule=0.5pt,
    left=2mm, right=2mm, top=1mm, bottom=1mm,
]

\textbf{Input:} \texttt{<path d="M 20,40 L 50,10 L 80,40 L 20,40" />} \\
\textbf{Task:} Identify the shape represented by the SVG path. \\
\textbf{Options:} (A) circle (B) heptagon (C) hexagon (D) kite (E) line (F) octagon (G) pentagon (H) rectangle (I) sector (J) triangle \\[1em]

\textbf{Step 1: Parse and Map the SVG Path Commands} \\
\textbf{Result \correct:} The path contains commands \texttt{M}, \texttt{L}, \texttt{L}, \texttt{L}, indicating a sequence of straight-line segments. The point sequence (20,40) → (50,10) → (80,40) → (20,40) suggests a continuous and closed path consisting of three connected segments. \\
\textbf{Refined Result \correct:} Same as above. No correction needed.\\[1em]

\textbf{Step 2: Identify Path Closure and Count Vertices/Edges} \\
\textbf{Result \wrong:} The model failed to detect that the final point returns to the starting point, and incorrectly assumed the path was open. As a result, it counted only two line segments and classified the shape as a line (\textbf{E}). \\
\textbf{Refined Result \correct:} After recognizing the coordinate repetition, the path was confirmed to be closed. The vertex count was corrected to three, and the shape reclassified as \textbf{J: triangle}. \textit{(based on \mistake{} and \prevention{})}\\[1em]

\textbf{Step 3: Analyze Curved or Straight Line Features} \\
\textbf{Result \correct:} All commands involved were \texttt{L}, which represent straight-line segments in SVG path notation. No \texttt{A} arc commands were found. Therefore, the model deduced that the shape lacks curvature and consists solely of linear edges. \\
\textbf{Refined Result \correct:} Same as above. No correction needed.\\[1em]

\textbf{Step 4: Validate Shape Properties Against Geometric Definitions} \\
\textbf{Result \correct:} With three edges, no arcs, and no additional symmetry, the structure aligns with a triangle. The model ruled out quadrilaterals or curved forms by verifying vertex count and shape simplicity. \\
\textbf{Refined Result \correct:} Same as above. No correction needed.\\[1em]

\textbf{Step 5: Apply a Process of Elimination to Confirm the Shape} \\
\textbf{Result \correct:} The model excluded all shapes requiring more than three vertices (e.g., polygon, rectangle) and those requiring curvature (e.g., ellipse, sector). Triangle remained the only viable candidate. \\
\textbf{Refined Result \correct:} Same as above. No correction needed.\\[1em]

\textbf{Final Option:} After refinement, the model selects \textbf{(J) triangle} as the final answer.\correct
\end{tcolorbox}
\caption{Structured reasoning on an SVG path with intermediate refinement. An error in Step~2—failing to detect path closure—was successfully corrected via a refinement mechanism driven by the identified mistake and its prevention. All other steps were executed correctly.}
\label{fig:svg_step_refine_full}
\end{figure*}

\end{document}